\pdfoutput=1

\documentclass[11pt,a4paper]{article}
\usepackage[hyperref]{emnlp2018}
\usepackage{times}
\usepackage{latexsym}
\usepackage{amssymb}
\usepackage{amsthm}
\usepackage{url}
\usepackage{lipsum}
\usepackage{subfigure}%
\usepackage{framed}
\usepackage{amsthm}
\usepackage{bm}
\usepackage{booktabs}
\usepackage{adjustbox}
\usepackage{todonotes}

\usepackage{amsmath}  
\DeclareMathOperator*{\argmax}{argmax}
\DeclareMathOperator*{\argmin}{argmin}
\usepackage{booktabs}
\usepackage{mathtools}
\usepackage{algorithm}
\usepackage{algorithmic}
\usepackage{graphicx}
\usepackage{enumitem}
\usepackage{multirow}
\usepackage{caption}
\usepackage[normalem]{ulem}
\usepackage{xcolor,colortbl, makecell}
\newcommand\customfont[1]{{\usefont{T1}{perm}{m}{n}#1}}

\aclfinalcopy
\newcommand{\datasetname}{{\small\customfont{Swag}}}
\newcommand{\titledatasetname}{\customfont{Swag}}

\newcommand{\datasetnamelong}{{\small\customfont{S}}ituations {\small\customfont{W}}ith {\small\customfont{A}}dversarial {\small\customfont{G}}enerations}

\title{
\titledatasetname: A Large-Scale Adversarial Dataset for\\ Grounded Commonsense Inference
}

\author{Rowan Zellers$^\spadesuit$ \: \: 
  Yonatan Bisk$^\spadesuit$ \: \:
  Roy Schwartz$^{\spadesuit\heartsuit}$ \: \:
  Yejin Choi$^{\spadesuit\heartsuit}$\\
  $^\spadesuit$Paul G. Allen School of Computer Science \& Engineering, University of Washington \\
  $^\heartsuit$Allen Institute for Artificial Intelligence\\
  {\tt \{rowanz,ybisk,roysch,yejin\}@cs.washington.edu}\\ 
  \vspace{1mm} \url{https://rowanzellers.com/swag}
  }

\date{}

\begin{document}
\maketitle
\thispagestyle{plain}
\begin{abstract}
Given a partial description like ``she opened the hood of the car,'' humans can reason about the situation and anticipate what might come next (``then, she examined the engine''). 
In this paper, we introduce the task of grounded commonsense inference, 
unifying natural language inference and commonsense reasoning.

We present \datasetname, a new dataset with 113k multiple choice questions about a rich spectrum of grounded situations. To address the recurring challenges of the annotation artifacts and human biases found in many existing datasets, we propose \emph{Adversarial Filtering} (AF), a novel procedure that constructs a de-biased dataset by iteratively training an ensemble of stylistic classifiers, and using them to filter the data. To account for the aggressive adversarial filtering, we use state-of-the-art language models to massively oversample a diverse set of potential counterfactuals. Empirical results demonstrate that while humans can solve the resulting inference problems with high accuracy (88\%), various competitive models struggle on our task. We provide comprehensive analysis that indicates significant opportunities for future research. 

\end{abstract}
\section{Introduction}
\label{sec:intro}


When we read a story, we bring to it a large body of implicit knowledge about the physical world. For instance, given the context ``on stage, a woman takes a seat at the piano,'' shown in Table~\ref{tab:examples}, we can 
easily infer 
what the situation might \emph{look} like: a woman is giving a piano performance, with a crowd watching her. 
We can furthermore infer her likely \emph{next} action: 
she will most likely set her fingers on the piano keys and start playing.

This type of {natural language inference} requires 
commonsense reasoning, substantially broadening the scope of prior work that focused primarily on linguistic entailment \cite{the_entailment_definition_rte_people_used}. 
Whereas the dominant 
entailment paradigm asks if two natural language sentences (the `premise' and the `hypothesis') describe the same set of possible worlds \cite{dagan2006pascal, bowman2015snli},
here we focus on whether a (multiple-choice) ending describes a possible (\emph{future}) world that can be anticipated from the situation described in the premise, even when it is not strictly entailed. 
Making such inference necessitates a rich understanding about everyday physical situations, 
including object affordances \cite{gibson1979ecological} and frame semantics \cite{baker1998berkeley}. 

\begin{table}[t!]
  \centering\small
  \begin{tabular}{@{} p{0.2cm}p{7.0cm} @{}}
  \toprule
\multicolumn{2}{l}{\parbox{7.0cm}{On stage, a woman takes a seat at the piano. She}} \\ \rule{0pt}{2ex}
  & a) sits on a bench as her sister plays with the doll. \\
  & b) smiles with someone as the music plays. \\  
  & c) is in the crowd, watching the dancers. \\ 
  & \textbf{d) nervously sets her fingers on the keys.} \\\midrule
\multicolumn{2}{l}{\parbox{7.0cm}{A girl is going across a set of monkey bars. She}} \\ \rule{0pt}{2ex}
  & a) jumps up across the monkey bars. \\
  & b) struggles onto the monkey bars to grab her head. \\
  & \textbf{c) gets to the end and stands on a wooden plank.} \\
  & d) jumps up and does a back flip. \\ \midrule
\multicolumn{2}{l}{\parbox{7.0cm}{
The woman is now blow drying the dog. The dog}} \\ \rule{0pt}{2ex}
  & \textbf{a) is placed in the kennel next to a woman's feet.} \\
  & b) washes her face with the shampoo. \\
  & c) walks into frame and walks towards the dog. \\
  & d) tried to cut her face, so she is trying to do something very close to her face. 
  \\
    \bottomrule
  \end{tabular}
  \caption{Examples from \datasetname; the correct answer is {\bf bolded}. Adversarial Filtering ensures that stylistic models find all options equally appealing.}
  \label{tab:examples}
\end{table}

A first step toward grounded commonsense inference with today's deep learning machinery is to create a large-scale dataset. 
However, recent work has shown that human-written datasets are susceptible to \emph{annotation artifacts}: unintended stylistic patterns that give out clues for the gold labels \cite{gururangan2018annotation,poliak_hypothesis_2018}. 
As a result, models trained on such datasets with human biases run the risk of 
over-estimating the actual performance on the underlying task, 
and are vulnerable to adversarial or out-of-domain examples \cite{wang2018glue, glockner2018breaking}.



In this paper, we introduce 
Adversarial Filtering (AF), 
a new method to automatically detect and reduce stylistic artifacts.
We use this method to construct 
\datasetname:
an adversarial dataset with 113k multiple-choice questions. 
We start with pairs of temporally adjacent video captions, each with a context and a follow-up event that we \emph{know} is physically possible. 
We then use a state-of-the-art language model fine-tuned on this data to massively oversample a diverse set of possible negative sentence endings (or \emph{counterfactuals}). Next, we filter these candidate endings aggressively and adversarially using a committee of trained models to obtain a population of de-biased endings with similar stylistic features to the real ones.
Finally, these filtered counterfactuals are validated by crowd workers to further ensure data quality.

Extensive empirical results demonstrate unique contributions of our dataset, complementing existing datasets for natural langauge inference (NLI) \cite{bowman2015snli, williams17multisnli} and commonsense reasoning \cite{roemmele_choice_2011,mostafazadeh_corpus_2016,zhang_ordinal_2017}.
\textbf{First}, our dataset poses a new challenge of grounded commonsense inference that is easy for humans (88\%) while hard for current state-of-the-art NLI models (${<}60\%$). 
\textbf{Second}, our proposed adversarial filtering methodology allows for cost-effective construction of a large-scale dataset while substantially reducing known annotation artifacts. 
The generality of adversarial filtering allows it to be applied to build future datasets, ensuring that they serve as reliable benchmarks.

\section{\datasetname: Our new dataset}
\label{sec:overview}

\begin{figure}
\includegraphics[width=\linewidth]{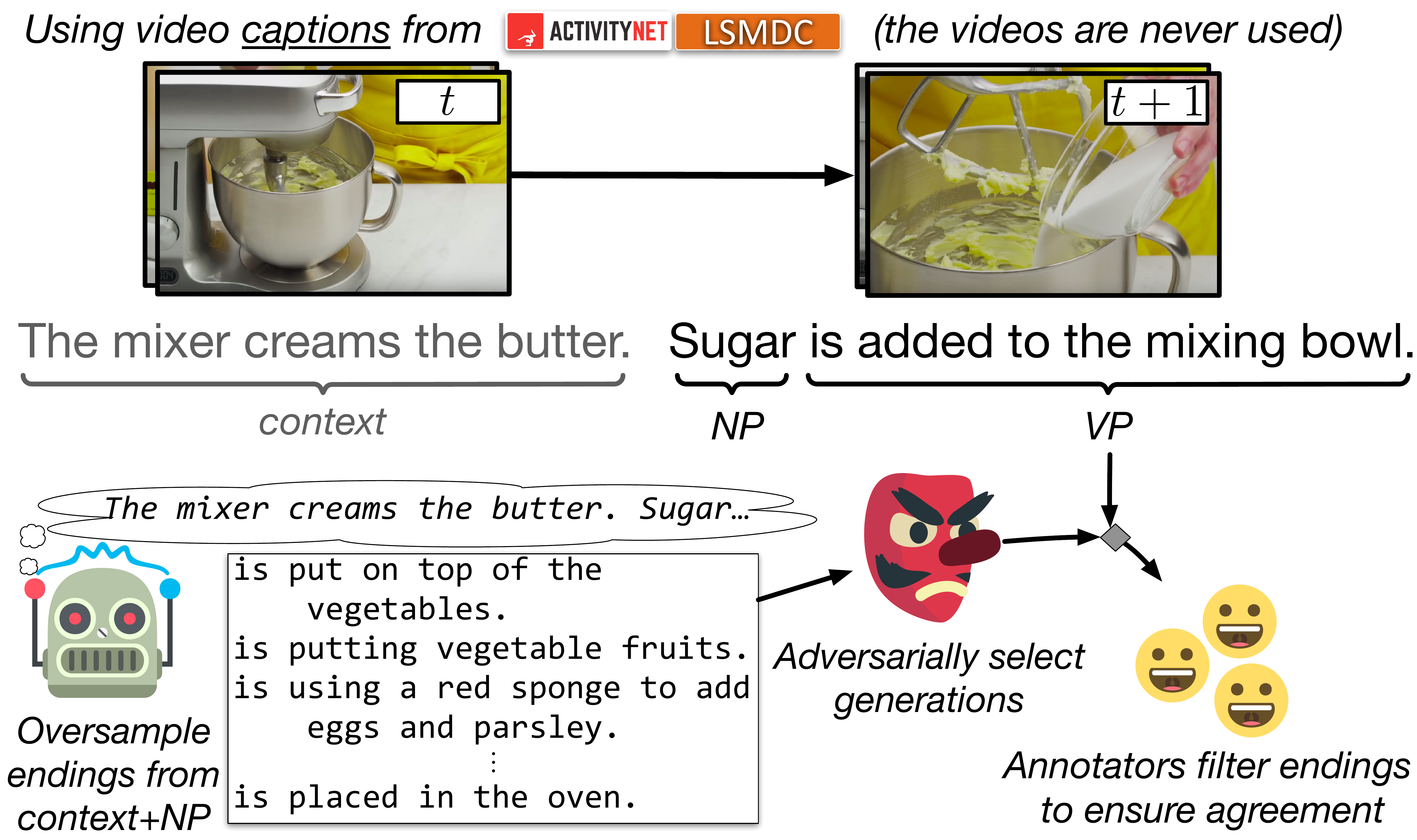}
\caption{Overview of the data collection process. For a pair of sequential video captions, the second caption is split into noun and verb phrases. A language model generates many negative endings, of which a difficult subset are human-annotated.
}
  \label{fig:teaser}
\end{figure}

We introduce a new dataset for studying physically grounded commonsense inference, called \datasetname.\footnote{Short for \datasetnamelong.} Our task is to predict which event is most likely to occur next in a video. More formally, a model is given a context $\bm{c} = (\bm{s}, \bm{n})$: a complete sentence $\bm{s}$ and a noun phrase $\bm{n}$ that begins a second sentence, as well as a list of possible verb phrase sentence endings $\bm{V} = \{\bm{v}_1, \ldots, \bm{v}_4\}$. See Figure~\ref{fig:teaser} for an example triple $(\bm{s}, \bm{n}, \bm{v}_i)$. The model must then select the most appropriate verb phrase $\bm{v}_{\hat{i}} \in \bm{V}$.

\paragraph{Overview} Our corpus consists of 113k multiple choice questions (73k training, 20k validation, 20k test) and is derived from pairs of consecutive video captions from ActivityNet Captions \cite{krishna_dense-captioning_2017,caba2015activitynet} and the Large Scale Movie Description Challenge (LSMDC; \citealp{rohrbach_movie_2017}). The two datasets are slightly different in nature and allow us to achieve broader coverage: ActivityNet contains 20k YouTube clips containing one of 203 activity types (such as doing gymnastics or playing guitar); LSMDC consists of 128k movie captions (audio descriptions and scripts). For each pair of captions, we use a constituency parser \cite{stern2017minimal} to split the second sentence into noun and verb phrases (Figure~\ref{fig:teaser}).\footnote{We filter out sentences with rare tokens ($\le$3 occurrences), that are short ($l\le5$), or that lack a verb phrase.} Each question has a human-verified gold ending and 3 distractors. 

\section{A solution to annotation artifacts}
In this section, we outline the construction of \datasetname. We seek dataset diversity while minimizing \emph{annotation artifacts}, conditional stylistic patterns such as length and word-preference biases. For many NLI datasets, these biases have been shown to allow shallow models (e.g. bag-of-words) obtain artificially high performance. 

To avoid introducing easily ``gamed" patterns, we present Adversarial Filtering (AF), a generally-applicable treatment involving the iterative refinement of a set of assignments to increase the entropy under a chosen model family. We then discuss how we generate counterfactual endings, and finally, the models used for filtering.

\subsection{Formal definition}
In this section, we formalize what it means for a dataset to be \emph{adversarial}. Intuitively, we say that an adversarial dataset for a model $f$ is one on which $f$ will not generalize, even if evaluated on test data from the same distribution. 
More formally, let our input space be $\mathcal{X}$ and the label space be $\mathcal{Y}$. Our trainable classifier $f$, taking parameters $\theta$ is defined as $f_\theta : \mathcal{X} \to \mathbb{R}^{|\mathcal{Y}|}$. Let our dataset of size $N$ be defined as $\mathcal{D} = \{(x_i, y_i)\}_{1 \le i \le N}$,  and let the loss function over the dataset be $L(f_\theta, \mathcal{D})$. We say that a dataset is \emph{adversarial} with respect to $f$ if we expect high empirical error $I$ over all leave-one-out train/test splits \cite{vapnik_nature_2000}:

{\setlength{\abovedisplayskip}{-11pt}
\setlength{\belowdisplayskip}{0pt}
\setlength{\abovedisplayshortskip}{0pt}
\setlength{\belowdisplayshortskip}{0pt}
\begin{align}
I(\mathcal{D}, f) &= \frac{1}{N} \sum_{i=1}^{N} L(f_{\theta_i^\star}, \{(x_i, y_i)\}), \\
\textrm{where } \theta_i^\star &= \argmin_\theta L(f_{\theta},\mathcal{D} \setminus \{(x_i, y_i)\}),
\end{align}
}
with regularization terms omitted for simplicity.
\begin{algorithm}[t]
\caption{\small Adversarial filtering (AF) of negative samples. During our experiments, we set $N^{easy}=2$ for refining a population of $N^{-}=1023$ negative examples to $k=9$, and used a 80\%/20\% train/test split.}
\begin{algorithmic}
\label{alg:adversarialfiltering}
\WHILE{convergence not reached}
    \STATE{$\bullet$ Split the dataset $\mathcal{D}$ randomly up into training and testing portions $\mathcal{D}^{tr}$ and $\mathcal{D}^{te}$.}
    \STATE{$\bullet$ Optimize a model $f_\theta$ on $\mathcal{D}^{tr}$.}
    
    \FOR{index $i$ in $\mathcal{D}^{te}$}
        \STATE{$\bullet$ Identify easy indices: \\
        $\mathcal{A}_i^{easy} = \{j\in \mathcal{A}_i:f_\theta(x^{+}_{i}) > f_\theta(x^{-}_{i,j})\}$}
        \STATE{$\bullet$ Replace $N^{easy}$ easy indices $j \in \mathcal{A}_i^{easy}$ with adversarial indices $k \not\in \mathcal{A}_i$ satisfying $f_\theta(x^{-}_{i,k}) > f_\theta(x^{-}_{i,j})$.}
   \ENDFOR
  \ENDWHILE
\end{algorithmic}
\end{algorithm}
\vspace{-1mm}
\subsection{Adversarial filtering (AF) algorithm}
In this section, we outline an approach for generating an adversarial dataset $\mathcal{D}$, effectively maximizing empirical error $I$ with respect to a family of trainable classifiers $f$. Without loss of generality, we consider the situation where we have $N$ \emph{contexts}, each associated with a single positive example $(x_i^{+}, 1)\,{\in}\,\mathcal{X}\,{\times}\,\mathcal{Y}$, and a large population of context-specific negative examples $(x_{i,j}^{-}, 0)\,{\in}\,\mathcal{X} \,{\times}\,\mathcal{Y}$, where $1{\le}j{\le}N^{-}$ for each $i$. For instance, the negative examples could be incorrect relations in knowledge-base completion \cite{socher2013reasoning}, or all words in a dictionary for a single-word cloze task \cite{zweig2011microsoft}.

Our goal will be to filter the population of negative examples for each instance $i$ to a size of $k{\ll}N^{-}$. This will be captured by returning a set of \emph{assignments} $\mathcal{A}$, where for each instance the assignment will be a $k$-subset $\mathcal{A}_i = [1\ldots N^{-}]^{k}$. The filtered dataset will then be:
{\setlength{\abovedisplayskip}{4pt}
\setlength{\belowdisplayskip}{4pt}
\setlength{\abovedisplayshortskip}{0pt}
\setlength{\belowdisplayshortskip}{0pt}
\begin{equation}
\mathcal{D}^{AF} = \{(x_i, 1), \{(x_{i,j}^{-}, 0)\}_{j \in \mathcal{A}_i}\}_{1 \le i \le N}
\end{equation}
}
Unfortunately, optimizing $I(\mathcal{D}^{AF}, f)$ is difficult as $\mathcal{A}$ is global and non-differentiable. To address this, we present Algorithm~\ref{alg:adversarialfiltering}. On each iteration, we split the data into dummy `train' and `test' splits. We train a model $f$ on the training portion and obtain parameters $\theta$, then use the remaining test portion to reassign the indices of $\mathcal{A}$. For each context, we replace some number of `easy' negatives in $\mathcal{A}$ that $f_\theta$ classifies correctly with `adversarial' negatives outside of  $\mathcal{A}$ that $f_\theta$ misclassifies. 

This process can be thought of as increasing the overall entropy of the dataset: given a strong model $f_\theta$ that is compatible with a random subset of the data, we aim to ensure it cannot generalize to the held-out set. We repeat this for several iterations to reduce the generalization ability of the model family $f$ over arbitrary train/test splits.

\subsection{Generating candidate endings}
To generate counterfactuals for \datasetname, we use an LSTM \cite{Hochreiter:1997:LSM:1246443.1246450} language model (LM), conditioned on contexts from video captions. We first pretrain on BookCorpus \cite{moviebook}, then finetune on the video caption datasets. The architecture uses standard best practices and was validated on held-out perplexity of the video caption datasets; details are in the appendix. We use the LM to sample $N^{-}{=}1023$ unique endings for a partial caption.\footnote{To ensure that the LM generates unique endings, we split the data into five validation folds and train five separate LMs, one for each set of training folds. This means that each LM never sees the found endings during training.} 

Importantly, we \emph{greedily} sample the endings, since beam search decoding biases the generated endings to be of lower perplexity (and thus easily distinguishable from found endings). We find this process gives good counterfactuals: the generated endings tend to use \emph{topical} words, but often make little sense physically, making them perfect for our task. Further, the generated endings are marked as ``gibberish'' by humans only 9.1\% of the time (Sec~\ref{subsec:human}); in that case the ending is filtered out. 

\subsection{Stylistic models for adversarial filtering}

\begin{figure}
    \centering
    \includegraphics[width=\linewidth]{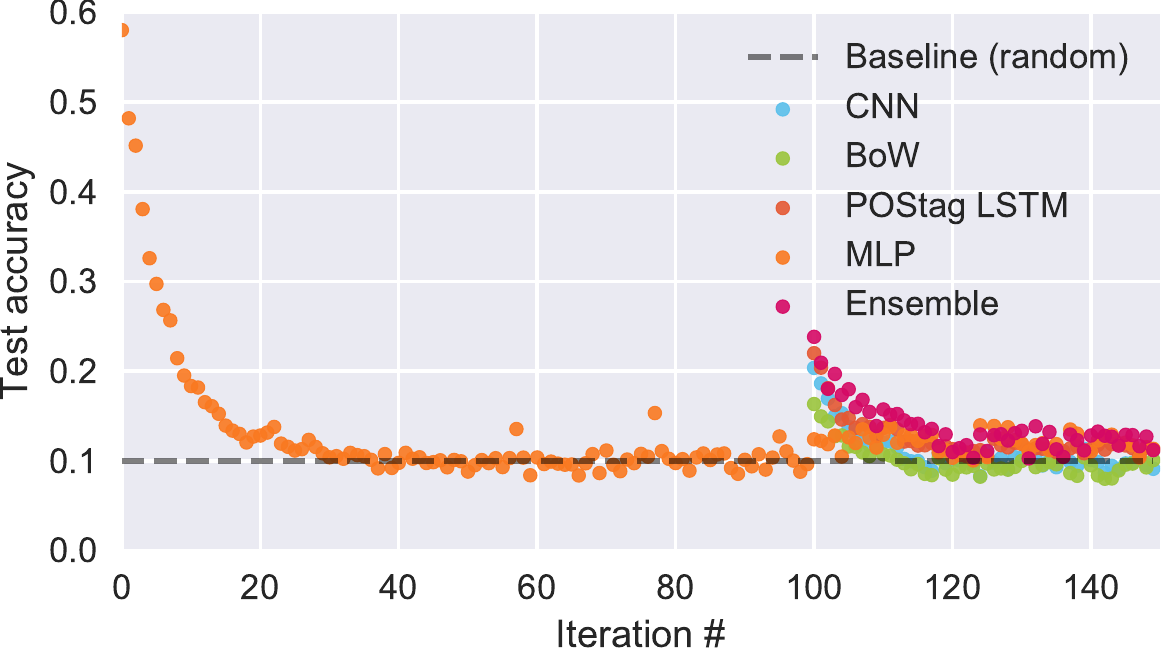}
    \vspace*{-7mm}
    \caption{Test accuracy by AF iteration, under the negatives given by $\mathcal{A}$.  
    The accuracy drops from around 60\% to close to random chance. For efficiency, the first 100 iterations only use the MLP.}\vspace{-1mm}
    \label{fig:adv_filtering}
\end{figure}
In creating \datasetname, we designed the model family $f$ to pick up on low-level \emph{stylistic features} that we posit should not be predictive of whether an event happens next in a video. These stylistic features are an obvious case of annotation artifacts \cite{cai2017pay, Schwartz:2017}.\footnote{A broad definition of annotation artifacts might include aspects besides lexical/stylistic features: for instance, certain events are less likely semantically regardless of the context (e.g. riding a horse using a hose). For this work, we erred more conservatively and only filtered based on style.} 
Our final classifier is an ensemble of four stylistic models:
\begin{enumerate}[wide, labelwidth=!,labelindent=0pt,noitemsep,topsep=0pt,label=\textbf{\arabic*}.]
    \item A multilayer perceptron (MLP) given LM perplexity features and context/ending lengths. 
    \item A bag-of-words model that averages the word embeddings of the second sentence as features. 
    \item A one-layer CNN, with filter sizes ranging from 2-5, over the second sentence. 
    \item A bidirectional LSTM over the 100 most common words in the second sentence; uncommon words are replaced by their POS tags. 
\end{enumerate}
We ensemble the models by concatenating their final representations and passing it through an MLP. On every adversarial iteration, the ensemble is trained jointly to minimize cross-entropy. 

The accuracies of these models (at each iteration, evaluated on a 20\% split of the test dataset before indices of $\mathcal{A}$ get remapped) are shown in Figure~\ref{fig:adv_filtering}. Performance decreases from 60\% to close to random chance; moreover, confusing the perplexity-based MLP is not sufficient to lower performance of the ensemble. Only once the other stylistic models are added does the ensemble accuracy drop substantially, suggesting that our approach is effective at reducing stylistic artifacts. 
\subsection{Human verification}\label{subsec:human}
\begin{figure}
    {\FrameSep6pt
    \begin{framed}
    \small
    Imagine that you are watching a video clip. The clip has a caption, but it is missing the final phrase. Please choose the best 2 caption endings, and classify each as:
    \begin{itemize}[wide, labelwidth=!,labelindent=0pt,noitemsep,topsep=2pt]
    \item \textbf{likely}, if it completes the caption in a reasonable way;
    \item \textbf{unlikely}, if it sounds ridiculous or impossible;
    \item \textbf{gibberish} if it has such serious errors that it doesn't feel like a valid English sentence.
    \end{itemize}
\vspace{2pt}
\emph{Example: Someone is shown sitting on a fence and talking to the camera while pointing out horses. Someone}
\begin{itemize}[wide, labelwidth=!,labelindent=0pt,noitemsep,topsep=2pt]
    \item stands in front of a podium. (\textbf{likely}, \textbf{second best})
    \item rides a horse using a hose. (\textbf{unlikely})
    \item is shown riding a horse. (\textbf{likely}, \textbf{best})
    \item , the horse in a plaza field. (\textbf{gibberish})
\end{itemize}
    \end{framed}}
    \vspace*{-3mm}\caption{Mechanical Turk instructions (abridged).}\vspace{-1mm}
    \label{fig:instructions}
\end{figure}

The final data-collection step is to have humans verify the data. Workers on Amazon Mechanical Turk were given the caption context, as well as six candidate endings: one found ending and five adversarially-sampled endings. The task was twofold: Turkers ranked the endings independently as \underline{likely}, \underline{unlikely}, or \underline{gibberish}, and 
selected the 
\underline{best} and \underline{second best} endings (Fig~\ref{fig:instructions}). 

We obtained the correct answers to each context in two ways. If a Turker ranks the found ending as either best or second best (73.7\% of the time), we add the found ending as a gold example, with negatives from the generations not labelled \underline{best} or \underline{gibberish}. Further, if a Turker ranks a generated ending as \underline{best}, and the found ending as \underline{second best}, then we have reason to believe that the generation is good. This lets us add an additional training example, consisting of the generated \underline{best} ending as the gold, and remaining generations as negatives.\footnote{These two examples share contexts. To prevent biasing the test and validation sets, we didn't perform this procedure on answers from the evaluation sets' context.} Examples with ${\le}3$ non-gibberish endings were filtered out.\footnote{To be data-efficient, we reannotated filtered-out examples by replacing \underline{gibberish} endings, as well as generations that outranked the found ending, with candidates from $\mathcal{A}$.}

\begin{table}[t!]
\centering\small
\begin{tabular}{@{}r | c c |c c @{}}
 & \multicolumn{2}{c|}{\parbox{2.8cm}{Label distribution by ending type}} & \multicolumn{2}{c}{\parbox{2cm}{Inter-annotator agreement}} \rule{0pt}{2ex} \\ \hline \rule{0pt}{2ex}
Labels & Found end & Gen. end & $\alpha$ & ppa\\ \hline \rule{0pt}{2ex}
Best & 53.5\% & 9.3\% & \multirow{3}{*}{0.43} & \multirow{3}{*}{72\%}\\
Second Best & 20.2\% & 15.9\% &\\
Neither & 26.3\% & 74.8\%& \\ \hline \rule{0pt}{2ex}
Likely & 80.3\% & 33.3\% & \multirow{3}{*}{0.39} & \multirow{3}{*}{64\%}\\
Unlikely & 19.0\% & 57.5\% \\
Gibberish & 0.7\% & 9.1\% \\ \hline
\end{tabular}
\vspace{-2mm}
\caption{Annotators tend to label the found ending as \underline{likely} and within the top 2 (column 2), in other cases the example is filtered out. Both label groups have high inter-annotator agreement, in terms of Krippendorff's $\alpha$ and pairwise percent agreement.
}
\label{tab:datastats3}
\end{table}
We found after 1000 examples that the annotators tended to have high agreement, also generally choosing found endings over generations (see Table~\ref{tab:datastats3}). Thus, we collected the remaining 112k examples with one annotator each, periodically verifying that annotators preferred the found endings.

\section{Experiments}
In this section, we evaluate the performance of various NLI models on \datasetname. Recall that models for our dataset take the following form: given a sentence and a noun phrase as context $\bm{c} = (\bm{s}, \bm{n})$, as well as a list of possible verb phrase endings $\bm{V} = \{\bm{v}_1, \ldots, \bm{v}_4\}$, a model $f_\theta$ must select a verb $\hat{i}$ that hopefully matches $i_{gold}$:
{\setlength{\abovedisplayskip}{5pt}
\setlength{\belowdisplayskip}{5pt}
\setlength{\abovedisplayshortskip}{0pt}
\setlength{\belowdisplayshortskip}{0pt}
    \begin{equation}
\hat{i} = \argmax_i f_\theta(\bm{s}, \bm{n}, \bm{v}_i)
\end{equation}
}
To study the amount of bias in our dataset, we also consider models that take as input just the ending verb phrase $\bm{v}_i$, or the entire second sentence $(\bm{n}, \bm{v}_i)$. For our learned models, we train $f$ by minimizing multi-class cross-entropy. We consider three different types of word representations: 300d GloVe vectors from Common Crawl \cite{pennington2014glove}, 300d Numberbatch vectors retrofitted using ConceptNet relations \cite{speer2017conceptnet}, and 1024d ELMo contextual representations that show improvement on a variety of NLP tasks, including standard NLI \cite{peters2018deep}. We follow the final dataset split (see Section~\ref{sec:overview}) using two training approaches: training on the found data, and the found and highly-ranked generated data. See the appendix for more details.

\subsection{Unary models}
The following models predict labels from \emph{a single span} of text as input; this could be the ending only, the second sentence only, or the full passage.

\begin{enumerate}[wide, labelwidth=!,labelindent=0pt,noitemsep,topsep=0pt,label=\textbf{\alph*}.]
\item {\bf fastText} \cite{joulin2017bag}: This library models a single span of text as a bag of $n$-grams, and tries to predict the probability of an ending being correct or incorrect independently.\footnote{The fastText model is trained using binary cross-entropy; at test time we extract the prediction by selecting the ending with the highest positive likelihood under the model.}
\item {\bf Pretrained sentence encoders}  We consider two types of pretrained RNN sentence encoders,  SkipThoughts \cite{kiros2015skip} and InferSent \cite{conneau2017supervised}. SkipThoughts was trained by predicting adjacent sentences in book data, whereas InferSent was trained on supervised NLI data. For each second sentence (or just the ending), we feed the encoding into an MLP.
\item {\bf LSTM sentence encoder} Given an arbitrary span of text, we run a two-layer BiLSTM over it. The final hidden states are then max-pooled to obtain a fixed-size representation, which is then used to predict the potential for that ending.
\end{enumerate}

\newcommand{\nodata}[1]{\multicolumn{#1}{|c|}{\cellcolor{gray!10}}}
\newcommand{\nodataright}[1]{\multicolumn{#1}{|c}{\cellcolor{gray!10}}}
\newcommand{\nd}{\cellcolor{gray!10}}

\newcommand{\tinyrule}{\cline{2-3} \\[-1.0em]}

\newcommand{\resultswidth}{1.35cm}
\begin{table*}[t!]
\centering
\begin{footnotesize}
\setlength{\tabcolsep}{4pt}
\begin{tabular}{@{} c@{\hspace{0.4em}}c@{\hspace{0.2em}} l @{\hspace{0.1em}}|
l@{\hspace{0.7em}}l@{\hspace{0.5em}} |l@{\hspace{0.7em}}l|l@{\hspace{0.7em}}l |l@{\hspace{0.7em}}l|l@{\hspace{0.7em}}l |l@{\hspace{0.7em}}l@{}}
& &\multicolumn{1}{c}{} & \multicolumn{4}{c}{Ending only} & \multicolumn{4}{c}{2nd sentence only} & \multicolumn{4}{c}{Context+2nd sentence}\\ 
& &\multicolumn{1}{c}{} & \multicolumn{2}{@{}c}{found only}& \multicolumn{2}{c}{found+gen}& \multicolumn{2}{c}{found only}& \multicolumn{2}{c}{found+gen}& \multicolumn{2}{c}{found only}& \multicolumn{2}{c@{}}{found+gen}\\
&  \multicolumn{2}{c|}{Model} & Val & Test & Val & Test & Val & Test & Val & Test & Val & Test & Val & Test \\ 
\toprule
&\multirow{3}{*}{\parbox{\resultswidth}{misc}}  & Random & 25.0 & 25.0 & 25.0 & 25.0 & 25.0 & 25.0 & 25.0 & 25.0 & 25.0 & 25.0 & 25.0 & 25.0 \\ 
& & Length & 26.7 & 27.0 & 26.7 & 27.0 & \nd&\nd&\nd&\nd&\nd&\nd&\nd&\nd \\ 
& & ConceptNet & \nd&\nd&\nd&\nd & 26.0 & 26.0 & 26.0 & 26.0 & \nd&\nd&\nd&\nd\\ 
\toprule
\multirow{6}{*}{\rotatebox[origin=c]{90}{Unary models}}&
& fastText & 27.5 & 26.9 & 29.9 & 29.0 & 29.2&27.8 & 29.8&29.0 & 29.4&28.0&30.3&29.8 \\ 
\tinyrule
& \multirow{2}{*}{\parbox{\resultswidth}{Sentence encoders}} & SkipThoughts & 32.4&32.1 &32.2 & 31.8& 33.0 & 32.4 & 32.8 & 32.3 & \nd&\nd&\nd&\nd \\
& & InferSent &30.6 & 30.2& 32.0& 31.9& 33.2 & 32.0 & 34.0 & 32.6 & \nd&\nd&\nd&\nd \\ 
\tinyrule
& \multirow{3}{*}{\parbox{\resultswidth}{LSTM sequence model}} & LSTM+GloVe & 31.9 & 31.8 & 32.9 & 32.4 & 32.7 & 32.4 & 34.3 & 33.5 & 43.1 & 43.6 & 45.6 & 45.7 \\
& & LSTM+Numberbatch & 32.4 & 32.6 & 32.3 & 31.9 & 31.9 & 31.9 & 34.1 & 32.8 & 39.9 & 40.2 & 41.2 & 40.5 \\
& & LSTM+ELMo & \bf{43.6} & \bf{42.9} & \bf{43.3} & \bf{42.3} & \bf{47.4} & \bf{46.7} & \bf{46.3} & \bf{46.0} & 51.4 & 50.6 & 51.3 & 50.4 \\ 
\toprule
\multirow{14}{*}{\rotatebox[origin=c]{90}{Binary models}} & \multirow{2}{*}{\parbox{\resultswidth}{DualBoW}} & DualBoW+GloVe & \nd&\nd&\nd&\nd & 31.3 & 31.3 & 31.9 & 31.2& 34.5 & 34.7 & 32.9 & 33.1 \\
& & DualBoW+Numberbatch & \nd&\nd&\nd&\nd & 31.9 & 31.4 & 31.6 & 31.3 & 35.1 & 35.1 & 34.2 & 34.1 \\ 
\tinyrule
& \multirow{4}{*}{\parbox{\resultswidth}{Dual \\sentence encoders}} & SkipThoughts-MLP &\nd&\nd&\nd&\nd& 34.6 & 33.9 & 36.2 & 35.5 & 33.4 & 32.3 & 37.4 & 36.4 \\
& & SkipThoughts-Bilinear &\nd&\nd&\nd&\nd & 36.0 & 35.7 & 34.7 & 34.5 & 36.5 & 35.6 & 35.3 & 34.9 \\
& & InferSent-MLP & \nd&\nd&\nd&\nd& 32.9 & 32.1 & 32.8 & 32.7 & 35.9 & 36.2 & 39.5 & 39.4 \\
& & InferSent-Bilinear &\nd&\nd&\nd&\nd& 32.0& 31.3 & 31.6 & 31.3& 40.5 & 40.3 & 39.0 & 38.4 \\ 
\tinyrule
& \multirow{2}{*}{\parbox{\resultswidth}{SNLI \\ inference}} & SNLI-ESIM  &\nd&\nd&\nd&\nd&\nd&\nd&\nd&\nd& 36.4 &36.1 &36.2 &36.0 \\
& & SNLI-DecompAttn  & \nd&\nd&\nd&\nd&\nd&\nd&\nd&\nd& 35.8 &35.8&35.8&35.7 \\
\tinyrule
& \multirow{6}{*}{\parbox{\resultswidth}{SNLI \\ models (retrained)}} & DecompAttn+GloVe & \nd&\nd&\nd&\nd&29.8 & 30.3 & 31.1 & 31.7 & 47.4 & 47.6 & 48.5 & 48.6 \\
& & DecompAttn+Numberbatch &  \nd&\nd&\nd&\nd&32.4 & 31.7 & 32.5 & 31.9&47.4 & 48.0 & 48.0 & 48.3 \\
& & DecompAttn+ELMo  & \nd&\nd&\nd&\nd& 43.4 & 43.4 & 40.6 & 40.3 & 47.7 & 47.3 & 46.0 & 45.4 \\ 
& & ESIM+GloVe &  \nd&\nd&\nd&\nd& 34.8 & 35.1 & 36.3 & 36.7& 51.9 & 52.7 & 52.5 & 52.5 \\
& & ESIM+Numberbatch & \nd&\nd&\nd&\nd& 33.1 & 32.6 & 33.0 & 32.4& 46.5 & 46.4 & 44.0 & 44.6 \\
& & ESIM+ELMo  &  \nd&\nd&\nd&\nd&46.0 & 45.7 & 45.9 & 44.8 & \bf{59.1} & \bf{59.2} & \bf{58.7} & \bf{58.5} \\ 
\toprule
& \multirow{4}{*}{\parbox{\resultswidth}{Human}} & 1 turker & \nd&\nd&\nd&\nd&\nd&\nd&\nd&\nd& \multicolumn{4}{c}{82.8} \\
& & 3 turkers &\nd&\nd&\nd&\nd&\nd&\nd&\nd&\nd&  \multicolumn{4}{c}{85.1} \\
& & 5 turkers &\nd&\nd&\nd&\nd&\nd&\nd&\nd&\nd& \multicolumn{4}{c}{\bf{88.0}} \\
& & Expert &\nd&\nd&\nd&\nd&\nd&\nd&\nd&\nd&  \multicolumn{4}{c}{85.0} \\ 
\bottomrule
\end{tabular}
\end{footnotesize}
\vspace*{-2mm}\caption{Performance of all models in accuracy (\%). All models substantially underperform humans, although performance increases as more context is provided (left to right).
We optionally train on \underline{found} endings {only}, or found and human-validated generated endings (\underline{found+gen}).
}\vspace*{-2mm}
\label{tab:results}
\end{table*}

\subsection{Binary models}
The following models predict labels from \emph{two spans} of text. We consider two possibilties for these models: using just the second sentence, where the two text spans are $\bm{n}, \bm{v}_i$, or using the context and the second sentence, in which case the spans are $\bm{s}, (\bm{n},\bm{v}_i)$. The latter case includes many models developed for the NLI task.

\begin{enumerate}[wide, labelwidth=!,labelindent=0pt,noitemsep,topsep=0pt,label=\textbf{\alph*}.]
\setcounter{enumi}{3}
\item {\bf Dual Bag-of-Words} For this baseline, we treat each sentence as a bag-of-embeddings $(\mathbf{c}, \mathbf{v}_i)$. We model the probability of picking an ending $i$ using a bilinear model: $\textrm{softmax}_i(\mathbf{c}\mathbf{W}\mathbf{v}_i^T)$.\footnote{We also tried using an MLP, but got worse results.}
\item {\bf Dual pretrained sentence encoders} Here, we obtain representations from SkipThoughts or InferSent for each span, and compute their pairwise compatibility using either 1) a bilinear model or 2) an MLP from their concatenated representations.
\item {\bf SNLI inference} Here, we consider two models that do well on SNLI \cite{bowman2015snli}: Decomposable Attention \cite{parikh2016decomposable} and ESIM \cite{chen2017enhanced}. 
We use pretrained versions of these models (with ELMo embeddings) on SNLI to obtain 3-way entailment, neutral, and contradiction probabilities for each example. We then train a log-linear model using these 3-way probabilities as features.
\item {\bf SNLI models (retrained)} Here, we train ESIM and Decomposable Attention on our dataset: we simply change the output layer size to 1 (the potential of an ending $\bm{v}_i$) with a softmax over $i$.
\end{enumerate}

\subsection{Other models}
We also considered the following models:
\begin{enumerate}[wide, labelwidth=!,labelindent=0pt,noitemsep,topsep=0pt,label=\textbf{\alph*}.]
\setcounter{enumi}{7}
\item {\bf Length}: Although length was used by the adversarial classifier, we want to verify that human validation didn't reintroduce a length bias. For this baseline, we always choose the shortest ending.
\item {\bf ConceptNet} As our task requires world knowledge, we tried a rule-based system on top of the ConceptNet knowledge base \cite{speer2017conceptnet}. For an ending sentence, we use the spaCy dependency parser to extract the head verb and its dependent object. The ending score is given by the number of ConceptNet causal relations\footnote{We used the relations `Causes', `CapableOf', `ReceivesAction', `UsedFor', and `HasSubevent'. Though their coverage is low (30.4\% of questions have an answer with $\ge$1 causal relation), the more frequent relations in ConceptNet, such as `IsA', at best only indirectly relate to our task.} between synonyms of the verb and synonyms of the object. 
\item {\bf Human performance} To benchmark human performance, 
five Mechanical Turk workers were asked to answer 100 dataset questions, as did an `expert' annotator (the first author of this paper). Predictions were combined using a majority vote. 
\end{enumerate}

\subsection{Results}
We present our results in Table~\ref{tab:results}. The best model that only uses the ending is the LSTM sequence model with ELMo embeddings, which obtains 43.6\%. This model, as with most models studied, greatly improves with more context: by 3.1\% when given the initial noun phrase, and by an additional 4\% when also given the first sentence. 

Further improvement is gained from models that compute pairwise representations of the inputs. While the simplest such model, DualBoW, obtains only 35.1\% accuracy, combining InferSent sentence representations gives 40.5\% accuracy (InferSent-Bilinear). The best results come from pairwise NLI models: when fully trained on \datasetname, ESIM+ELMo obtains 59.2\% accuracy.

When comparing machine results to human results, we see there exists a lot of headroom. Though there likely is some noise in the task, our results suggest that humans (even untrained) converge to a consensus. Our in-house ``expert'' annotator is outperformed by an ensemble of 5 Turk workers (with 88\% accuracy); thus, the effective upper bound on our dataset is likely even higher.

\section{Analysis}
\begin{figure*}[t]
\centering\includegraphics[width=\linewidth]{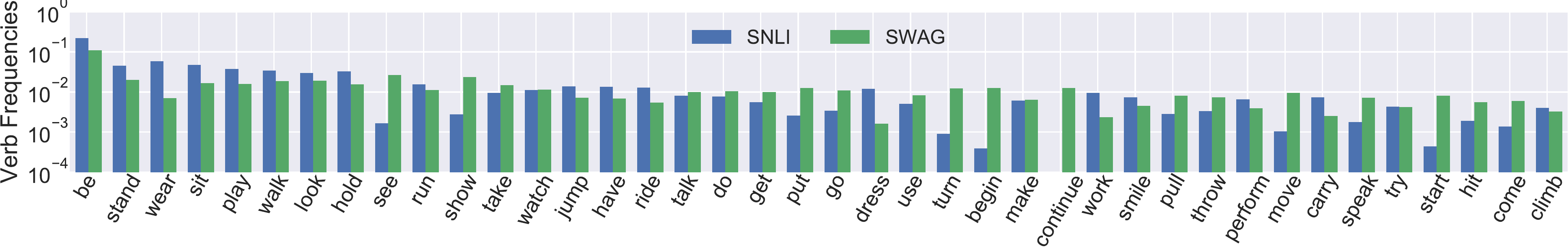}
\vspace{-2mm}
\centering\includegraphics[width=\linewidth]{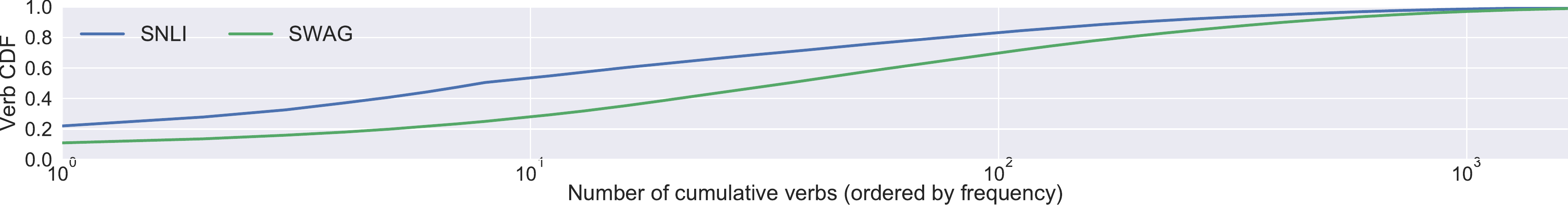}
    \caption{Top: Distribution of the 40 top verbs in the union of SNLI and \datasetname. Our dataset shows a greater variety of dynamic verbs, such as ``move'', as well as temporal verbs such as ``start'' and ``come.'' ``Continue'' is cut off for SNLI (it has frequency $6\cdot 10^{-5}$). Bottom: CDF for verbs in SNLI and \datasetname.
    }
    \label{fig:diversity}
\end{figure*}

\subsection{\datasetname~versus existing NLI datasets}
The past few years have yielded great advances in NLI and representation learning, due to the availability of large datasets like SNLI and MultiNLI \cite{bowman2015snli,williams17multisnli}. With the release of \datasetname, we hope to continue this trend, particularly as our dataset largely has the same input/output format as other NLI datasets. 
We observe three key differences between our dataset and others in this space:

First, as noted in Section~\ref{sec:intro}, \datasetname~requires a unique type of temporal reasoning. A state-of-the-art NLI model such as ESIM, when bottlenecked through the SNLI notion of entailment (SNLI-ESIM), only obtains 36.1\% accuracy.\footnote{The weights of SNLI-ESIM pick up primarily on entailment probability (0.59), as with neutral (0.46), while contradiction is negatively correlated (-.42).} This implies that these datasets necessitate different (and complementary) forms of reasoning.

Second, our use of videos results in wide coverage of dynamic and temporal situations
Compared with SNLI, with contexts from Flickr30K \cite{Plummer2017flickr30k} image captions, \datasetname~has more active verbs like `pull' and `hit,' and fewer static verbs like `sit' and `wear' (Figure~\ref{fig:diversity}).\footnote{Video data has other language differences; notably, character names in LSMDC were replaced by `someone'}

Third, our dataset suffers from few lexical biases. Whereas fastText, a bag of $n$-gram model, obtains 67.0\% accuracy on SNLI versus a 34.3\% baseline \cite{gururangan2018annotation}, fastText obtains only 29.0\% accuracy on \datasetname.\footnote{The most predictive individual words on SWAG are infrequent in number: `dotted` with P$(+|\textrm{dotted})=77\%$ with 10.3 counts, and P$(-|\textrm{similar})=81\%$ with 16.3 counts. (Counts from negative endings were discounted 3x, as there are 3 times as many negative endings as positive endings).}

\begin{table}[t]
    \small
    \centering
    \begin{tabular}{@{}l @{\hspace{0.2cm}} p{4.8cm} @{\hspace{0.2cm}} l @{}}
        \toprule
        Reason & Explanation & Freq. \\
        \midrule
        \cmidrule{1-3}
         Situational & The good ending is better \emph{in context}. & 53.7\% \\
         Plausibility & The bad ending is implausible \emph{regardless of context}. & 14.4\% \\
         Novelty & The bad ending seems redundant; it is entailed by the context. & 1.8\% \\
         Weirdness & The bad ending is semantically or grammatically malformed, e.g. `the man is getting out of the horse.' & 18.1\% \\
         Ambiguous & Both endings seem equally likely. & 12.0\% \\
         \bottomrule
    \end{tabular}
    \vspace*{-2mm}\caption{Justifications for ranking the gold answer over a wrong answer chosen by ESIM+ELMo.}
    \label{tab:piechart}
\end{table}

\subsection{Error analysis}
We sought to quantify how human judgments differ from the best studied model, ESIM+ELMo. We randomly sampled 100 validation questions that ESIM+ELMo answered incorrectly, for each extracting both the gold ending and the model's preferred ending. We asked 5 Amazon Mechanical Turk workers to pick the better ending (of which they preferred the gold endings 94\% of the time) and to select one (or more) multiple choice reasons explaining why the chosen answer was better.

The options, and the frequencies, are outlined in Table~\ref{tab:piechart}. The most common reason for the turkers preferring the correct answer is \underline{situational} (52.3\% of the time), followed by \underline{weirdness} (17.5\%) and \underline{plausibility} (14.4\%). This suggests that ESIM+ELMo already does a good job at filtering out \underline{weird} and \underline{implausible} answers, with the main bottleneck being grounded physical understanding. The ambiguous percentage is also relatively low (12.0\%), implying significant headroom.
\newcommand{\aquestion}[5]{\begin{tabular}{p{0.2cm}p{6.9cm}}
\multicolumn{2}{l}{\parbox{7.0cm}{#1}} \\ 
& #2\\ 
& #3\\
& #4\\
& #5\\
\end{tabular}
}

\newcommand\mywidth{6.9cm}
\newcommand{\correctans}[1]{\textcolor{blue}{\textbf{#1}}}
\newcommand{\incans}[1]{\textcolor{red}{#1}}

\begin{table*}[!ht]
\centering\small
\begin{tabular}{@{}l @{\hspace{0.1cm}}|@{\hspace{0.1cm}} l@{}}
\aquestion{A waiter brings a fork. The waiter}{\correctans{a) starts to step away. (74.76\%)}}{b) adds spaghetti to the table. (21.57\%)}{c) brings a bunch of pie to the food (2.67\%)}{d) drinks from the mug in the bowl. (0.98\%)}
& 
\aquestion{He is up a tree. Someone}{\correctans{a) stands underneath the tree. (97.44\%)}}
{b) is at a pool table holding a cup. (1.14\%)}{c) grabs a flower from a paper. (0.96\%)}{d) is eating some cereal. (0.45\%)} 

\\ \midrule\rule{0pt}{2ex}
\aquestion{An old man rides a small bumper car. Several people}{
\incans{a) get in the parking lot. (76.58\%)}}{b) wait in the car. (15.28\%)}{\textbf{c) get stuck with other bumper cars. (6.75\%)}}{d) are running down the road. (1.39\%)} 
& 
\aquestion{He pours the raw egg batter into the pan. He}{\incans{a) drops the tiny pan onto a plate. (93.48\%)}}{\textbf{b) lifts the pan and moves it around to shuffle the eggs. (4.94\%)}}{c) stirs the dough into a kite. (1.53\%)}{d) swirls the stir under the adhesive. (0.05\%)}
\\ \midrule
\end{tabular}
\vspace*{-3mm}\caption{Example questions answered by the best model, ESIM+Elmo, sorted by model probability. Correct model predictions are in \correctans{blue}, incorrect model predictions are \incans{red}. The right answers are \textbf{bolded}.}
\label{tab:qualitative}
\end{table*}

\subsection{Qualitative examples}
Last, we show several qualitative examples in Table~\ref{tab:qualitative}. Though models can do decently well by identifying complex alignment patterns between the two sentences (e.g. being ``up a tree'' implies that ``tree'' is the end phrase), the incorrect model predictions suggest this strategy is insufficient. For instance, answering ``An old man rides a small bumper car'' requires knowledge about \emph{bumper cars} and how they differ from regular cars: bumper cars are tiny, don't drive on roads, and don't work in parking lots, eliminating the alternatives. However, this knowledge is difficult to extract from existing corpora: for instance, the ConceptNet entry for Bumper Car has only a single relation: bumper cars are a type of vehicle. Other questions require intuitive physical reasoning: e.g, for ``he pours the raw egg batter into the pan,'' about what happens next in making an omelet.

\subsection{Where to go next?}
Our results suggest that \datasetname~is a challenging testbed for NLI models. However, the adversarial models used to filter the dataset are purely stylistic and focus on the second sentence; thus, subtle artifacts still likely remain in our dataset. These patterns are ostensibly picked up by the NLI models (particularly when using ELMo features), but the large gap between machine and human performance suggests that more is required to solve the dataset. As models are developed for commonsense inference, and more broadly as the field of NLP advances, we note that AF can be used again to create a more adversarial version of \datasetname~using better language models and AF models.

\section{Related Work}
\paragraph{Entailment NLI} 
There has been a long history of NLI benchmarks focusing on linguistic entailment \cite{cooper1996framework, dagan2006pascal, marelli2014sick,bowman2015snli, Lai:2017, williams17multisnli}. 
Recent NLI datasets in particular have supported learning broadly-applicable sentence representations \cite{conneau2017supervised}; moreover, models trained on these datasets were used as components for performing better 
video captioning \cite{pasunuru2017multitask}, summarization \cite{pasunuru_multireward_2018}, and generation \cite{holtzman2018learning}, confirming the importance of NLI research. 
The NLI task requires a variety of commonsense knowledge \cite{lobue2011types}, which our work complements. However, previous datasets for NLI have been challenged by unwanted annotation artifacts, \cite{gururangan2018annotation,poliak_hypothesis_2018} or scale issues. Our work addresses these challenges by constructing a new NLI benchmark focused on grounded commonsense reasoning, and by introducing an adversarial filtering mechanism that substantially reduces known and easily detectable annotation artifacts.

%

\paragraph{Commonsense NLI} Several datasets have been introduced to study NLI beyond linguistic entailment: for inferring likely causes and endings given a sentence (COPA; \citealp{roemmele_choice_2011}), for choosing the most sensible ending to a short story (RocStories; \citealp{mostafazadeh_corpus_2016,sharma2018tackling}), and for predicting likelihood of a hypothesis by regressing to an ordinal label (JOCI; \cite{zhang_ordinal_2017}). These datasets are relatively small: 1k examples for COPA and 10k cloze examples for RocStories.\footnote{For RocStories, this was by design to encourage learning from the larger corpus of 98k sensible stories.} JOCI increases the scale by generating the hypotheses using a knowledge graph or a neural model. 
In contrast to JOCI where the task was formulated as a regression task on the degree of plausibility of the hypothesis, we frame commonsense inference as a multiple choice question to reduce the potential ambiguity in the labels and to allow for direct comparison between machines and humans. 
In addition, \datasetname's use of adversarial filtering increases diversity of situations and counterfactual generation quality.

Last, another related task formulation is sentence completion or cloze, where the task is to predict a single word that is removed from a given context \cite{zweig2011microsoft,denis2016lambada}.\footnote{Prior work on sentence completion filtered negatives with heuristics based on LM perplexities. We initially tried something similar, but found the result to still be gameable.} 
Our work in contrast requires  longer textual descriptions to reason about. 

\paragraph{Vision datasets} Several resources have been introduced to study temporal inference in vision. The Visual Madlibs dataset has 20k image captions about hypothetical next/previous events \cite{yu_visual_2015}; similar to our work, the test portion is multiple-choice, with counterfactual answers retrieved from similar images and verified by humans. The question of `what will happen next?' has also been studied in photo albums \cite{huang2016visual}, videos of team sports, \cite{felsen2017will} and egocentric dog videos \cite{Ehsani_2018_CVPR}. Last, annotation artifacts are also a recurring problem for vision datasets such as Visual Genome \cite{zellers2018scenegraphs} and Visual QA \cite{jabri2016revisiting}; recent work was done to create a more challenging VQA dataset by annotating complementary image pairs \cite{goyal2017making}.

\paragraph{Reducing gender/racial bias} Prior work has sought to reduce demographic biases in word embeddings \cite{zhang2018mitigating} as well as in image recognition models \cite{zhao2017men}. Our work has focused on producing a dataset with minimal annotation artifacts, which in turn helps to avoid some gender and racial biases that stem from elicitation \cite{rudinger2017social}. However, it is not perfect in this regard, particularly due to biases in movies \cite{schofield2016gender, sap2017connotation}. Our methodology could potentially be extended to construct datasets free of (possibly intersectional) gender or racial bias. 

\paragraph{Physical knowledge} Prior work has studied learning grounded knowledge about objects and verbs: from knowledge bases \cite{li_commonsense_2016}, syntax parses \cite{forbes2017verb}, word embeddings \cite{lucy2017distributional}, and images and dictionary definitions \cite{emnlp17_zellers}. An alternate thread of work has been to learn scripts: high-level representations of event chains \cite{Schank1975,chambers_unsupervised_2009}. \datasetname~evaluates both of these strands.

\section{Conclusion}
We propose a new challenge of physically situated commonsense inference that broadens the scope of natural language inference (NLI) with commonsense reasoning. To support research toward commonsense NLI, we create a large-scale dataset \datasetname\  with 113k multiple-choice questions. Our dataset is constructed using Adversarial Filtering (AF), a new paradigm for robust and cost-effective dataset construction that allows datasets to be constructed at scale while automatically reducing annotation artifacts that can be easily detected by a committee of strong baseline models. Our adversarial filtering paradigm is general, allowing potential applications to other datasets that require human composition of question answer pairs. 


\section*{Acknowledgements}
We thank the anonymous reviewers, members of the ARK and xlab at the University of Washington, researchers at the Allen Institute for AI, and Luke Zettlemoyer for their helpful feedback. We also thank the Mechanical Turk workers for doing a fantastic job with the human validation. This work was supported by the National Science Foundation Graduate Research Fellowship (DGE-1256082), the NSF grant (IIS-1524371, 1703166), the DARPA CwC program through ARO (W911NF-15-1-0543), the IARPA DIVA program through D17PC00343, and gifts by Google and Facebook. The views and conclusions contained herein are those of the authors and should not be interpreted as 
representing endorsements 
of IARPA, DOI/IBC, or the U.S. Government.

\appendix
\section{Appendix}

\subsection{More detail about video datasets}
As mentioned in the main paper, we obtained contexts and found endings from video data. The videos in the ActivityNet dataset are already broken up into into clips. However, the LSMDC dataset contains captions for the entire movie, so it is possible that temporally adjacent captions describe events that are far apart in time. Thus, we don't include any pair of captions that have a time-difference of more than 25 seconds.

In addition to the datasets we used, we also considered the DiDeMo dataset, which consists of (often several) referring expressions in a video \cite{hendricks17iccv}. However, many of the referring expressions are themselves sentence fragments, (e.g. ``first time we see people'' so we ultimately did not use this dataset.) Additionally, we considered the Visual Madlibs dataset \cite{yu_visual_2015}, as it contains 10k hypothetical captions written by Mechanical Turk workers about what might happen \emph{next} given an image. However, these captions are fundamentally different from the rest of the data (as they're about what \emph{might}) happen next; as a result, they use different types of language. They also have different tenses versus the other datasets that we considered (e.g. past tense), as a result of the ``Mad-libs'' style of data collection.

\subsection{Details of the language model}
 Our language model follows standard best practices: the input and output embedding layers are tied \citep{inan2016tying, press2017using}, all embedding and hidden layers are set to 512, and we used recurrent dropout \cite{gal2016theoretically} on the hidden states and embedding layer. We additionally train a \emph{backwards} language model alongside the forward language model, and they share embedding parameters. This adds extra supervision to the embedding layer and gives us another way to score candidate generations. We first pretrain the language model for two epochs on pairs of two sentences in the Toronto Books dataset \cite{moviebook}, and then train on sentence pairs from ActivityNet Captions and LSMDC, validating on held-out perplexity. For optimization, we use Adam \cite{Kingma2014AdamAM} with a learning rate of $10^{-3}$ and clip gradients to norm $1.0$. 
 
 All of the above details were validated using perplexity on a held-out set of the video datasets during early experimentation. Our final development set forward perplexity was 31.2 and backward perplexity was 30.4. We tried more complicated language modeling architectures, such as from \cite{Jzefowicz2016ExploringTL}, but ended up not seeing an improvement due to overfitting.

\subsection{Language model features for the MLP, during adversarial filtering}
We obtained LM perplexity features to be used during adversarial filtering in the following ways, using both directions of the bidirectional language model. We extract perplexities for the context by itself (going forward), the ending given the context (going forward), the context given the ending (going backward), and the ending by itself (going backward). We also extract the probability of the final generated token going forward, since sentences sometimes reach the length limit of 25 tokens and end unnaturally.

\subsection{Refinining the generated answers to four distractors}
In the main paper, we noted that we started with 1023 negatives per example, which the adversarial filtering process filtered down to 9. Five of these were passed to mechanical turk workers, and we were left with anywhere between 0 and 4 of these per example as ``distractors.'' (Note that we always were filtering out the \underline{second best} option that the was selected by the turkers). This means that for many of our examples (62\%) we actually have a fourth distractor. In these cases, we sorted the distractors by their ``unlikely/likely'' score, so that the fourth distractor was the one deemed most likely. We still provided the fourth distractor in the training set to be possibly used in future work, however we didn't train on it for simplicity.

\subsection{More information about Mechanical turk}
We used several tricks to keep the interannotator agreement high (with a pairwise percent agreement of 79\% at classifying an ending as either in the Top 2). First, we had a screening HIT where turkers were given detailed instructions for the task, and only the best-scoring turk workers qualified for the remaining HITs. Second, we periodically dequalified turkers that had a low agreement with the gold endings: any turk worker with an accuracy of less than 55\% of classifying the ``gold'' ending as the best or second best, over 10 or more HITs, had the qualification taken away. We also gave small bonuses to turkers with high accuracy. 

During our crowdsourcing, we tried to pay the Turkers a fair wage (median \$8.57 per hour) and they left positive comments for us on TurkOpticon and TurkerView. The total dataset cost was \$23,000, or an average of 20 cents per example.

\subsection{Implementation details of the models considered}\label{appendix:Implementation}
We implemented the neural models in PyTorch using the AllenNLP library \cite{gardner2018allennlp}. Our experiments use the Adam optimizer \cite{Kingma2014AdamAM}, with a learning rate of $10^{-3}$ and gradient clipping, except for Decomposable Attention and ESIM, where we use the AllenNLP default configurations. 

\begin{table}[t!]
\centering
\begin{tabular}{@{}l r @{}}
\toprule
Questions with only generated endings & 25,618 \\
Questions with one original ending & 87,939 \\
Questions in total & 113,557 \\
Sentence pairs from ActivityNet & 51,439 \\
Sentence pairs from LSMDC & 62,118 \\
Unique contexts & 92,221 \\
Unique endings & 452,683 \\
\bottomrule
\end{tabular}
\caption{Statistics of \datasetname.}
\label{tab:datastats}
\end{table}

\subsection{More info about dataset diversity}
The final dataset has a vocabulary size of 21000. We also visualize the coverage of the dataset with a Topic model (see Table~\ref{tab:lda}).

\begin{table}[t]
    \small
    \centering
    \begin{tabular}{@{}l @{\hspace{0.2cm}} c @{}}
        \toprule
        Freq & Topic words\\
        \midrule
        \cmidrule{1-2}
5.0\% & ball, pull, hit, wall, inside, time, game, rope, team \\ 
4.9\% & window, red, long, drink, bowl, 
ingredient, mix \\
6.1\% & arm, speak, appear, climb, tree, roll, like, roof, edge \\ 
4.0\% & water, bar, board, blue, boat, fly, 
river, join, dive \\ 
5.3\% & eye, smile, close, 
little,  lean, 
cover, remove, lip \\ 
4.6\% & walk, outside, street, wave, pass, 
beach, sidewalk \\ 
5.7\% & field, drop, slide, drive, right, kick, park, 
road, chest \\ 
4.7\% & watch, dog, flip, 
stick, land, 
demonstrate, trick, mat \\ 
4.5\% & dance, lift, try, line, snow, gun, catch, hill, bend \\ 
4.6\% & fall, crowd, pour, shake, finish, raise, 
grass, wooden \\ 
5.9\% & perform, spin, house, stage, 
routine, fence, bow \\ 
         \bottomrule
    \end{tabular}
    \caption{A visualization of the diversity of the dataset, using a topic model \cite{blei2003latent}. }
    \label{tab:lda}
\end{table}

\subsection{Comparing the distribution of verbs with MultiNLI}
\begin{figure*}[t]
\centering\includegraphics[width=\linewidth]{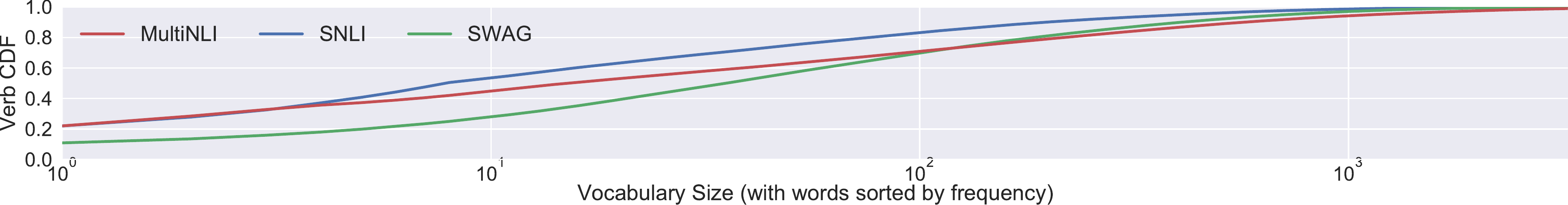}
    \caption{Bottom: CDF for verbs in SNLI, \datasetname, and MultiNLI.}
    \label{fig:diversityMULTINLI}
\end{figure*}

We also produced an extension to Figure 4 of the main paper, that involves verbs from MultiNLI, in Figure~\ref{fig:diversityMULTINLI}. We ended up not including it in the paper because we wanted to focus our comparison between SNLI and \datasetname~(as they are both grounded datasets). Interestingly, we find that \datasetname~has a less skewed cumulative distribution of verbs up to around 120, when afterwards MultiNLI has a slightly less skewed distribution. This is possibly due to the broader set of domains considered by MultiNLI, whereas we consider videos (which is also a broad domain! but still underrepresents words highly used in newswire text, for instance.)

\subsection{More examples}

We have more qualitative examples in Table~\ref{tab:morequalitative}.

\begin{table*}[!ht]
\centering\small
\begin{tabular}{@{}l @{\hspace{0.1cm}}|@{\hspace{0.1cm}} l@{}}
\aquestion{The lady demonstrates wrapping gifts using her feet. The lady}{\incans{a) shows us the different shapes of the ornaments. (99.67\%)}}{b) continues playing when the lady talks to the camera. (0.26\%)}{c) takes the desserts from the box and continues talking to the camera . (0.07\%)}{\textbf{d) cuts the paper with scissors. (0.01\%)}} &
\aquestion{In a cafeteria, someone holds a combination tray and bowl in one hand. With the other, he}{\incans{a) heads into his own study. (80.67\%)}}{b) glances around and studies the photo of the blonde someone. (8.45\%)}{\textbf{c) struggles to serve himself food with chopsticks. (6.82\%)}}{d) opens the wall , revealing an expanse of bed within. (4.06\%)}
\\ \midrule
\aquestion{As he approaches , his kayak flips upside-down. As the view follows him, we}{\incans{a) see silhouetted black clouds making him zoom out of the trees, catching smoke. (42.54\%)}}{b) drift over a busy city street , like down buildings on the tarmac. (41.41\%)}{c) find someone climbing into a tawny grave atop a road drawn among german soldiers. (13.73\%)}{\textbf{d) notice another man seated on the rocks to the right in red with a white helmet. (2.32\%)}}
&
\aquestion{A man is bending over a sink. He }{ \incans{a) takes a rag from over the sink, putting it in his mouth. (89.54\%)} }{ \textbf{ b) is spraying a small dog with a hose. (6.07\%) } }{ c) is carrying a shaving machine with a pressure washer. (4.29\%) }{ d) is putting a pair of shaving glass on the side of his face. (0.10\%) } \\ \midrule
\aquestion{People are walking next to the camels leading them. A building }{ \incans{a) is shown riding the camels. (90.72\%)} }{ \textbf{ b) is shown in the background. (8.39\%) } }{ c) with a rifle is leading them. (0.87\%) }{ d) is then shown for several clip. (0.01\%) } & 
\aquestion{A hockey game is in progress. two hockey players }{ \incans{a) walked together in the middle of a field. (48.11\%)} }{ b) walk past with a goal. (44.00\%) }{ c) sit around a rope watching the other team. (5.30\%) }{ \textbf{ d) ram into each other and begin fighting. (2.58\%) } } \\ \midrule
\aquestion{Meanwhile, someone parries another giant 's attacks. The giant }{ \incans{a) strikes a fight and thuds into someone as he rushes in, who briefly flees. (89.96\%)} }{ \textbf{ b) knocks someone 's sword out of his hand. (5.25\%) } }{ c) spins him across the bars. (4.55\%) }{ d) throws stick to the bat, dragging around. (0.24\%) } &
\aquestion{A lady pours ice in a glass. The lady }{ \incans{a) pours ice into the glass. (65.14\%)} }{ b) measures the contents of the glass. (33.56\%) }{ c) pours lemon mixture into a glass and pours liquids into asian juice. (0.87\%) }{ \textbf{ d) adds 3 liquors and lemon juice. (0.43\%) } } \\ \midrule
\aquestion{The stars emerge from behind the clouds. Someone }{ \incans{a) backs away from the windows of the clip, as lightning billows over the sky. (96.59\%)} }{ b) walks back across the room with nothing of his own. (1.82\%) }{ \textbf{ c) stands on his boat and looks at a deep orange and red sunset. (1.47\%) } }{ d) shoots the man 's shoulder sideways, but neither do anything for a few seconds. (0.12\%) } &
\aquestion{Someone stands waiting with the bridesmaids. Everyone }{ \incans{a) seems to be ecstatic. (78.33\%)} }{ \textbf{ b) looks around as someone walks down the aisle, arm-in-arm with someone 's uncle. (8.97\%) } }{ c) holds someone 's eyebrow. (8.84\%) }{ d) looks at her anxiously as someone walks and sits in his seat. (3.85\%) } \\ \bottomrule
\end{tabular}
\caption{More (incorrect) questions answered by the best model, ESIM+Elmo, sorted by model probability. The right answers are \textbf{bolded}.}
\label{tab:morequalitative}
\end{table*}

\bibliography{emnlp2018}
\bibliographystyle{acl_natbib_nourl}

\end{document}